# Social Media Writing Style Fingerprint

**Himank Yadav, Juliang Li**
Department of Computer Science and Engineering
Texas A&M University
{hyadav, juliang0705}@tamu.edu

## Abstract

We present our approach for computer-aided social media text authorship attribution based on recent advances in short text authorship verification. We use various natural language techniques to create word-level and character-level models that act as hidden layers to simulate a simple neural network. The choice of word-level and character-level models in each layer was informed through validation performance. The output layer of our system uses an unweighted majority vote vector to arrive at a conclusion. We also considered writing bias in social media posts while collecting our training dataset to increase system robustness. Our system achieved a precision, recall and F-measure of 0.82, 0.926 and 0.869 respectively.

## 1 Introduction

Humans have the cognitive ability to understand various writing styles and distinguish between writing samples from different authors. This also allows humans to tell whether or not a writing sample belongs to a given author after having read enough of their articles. We want to apply computer-aided authorship detection on shorter writing samples that we gathered through social media to intelligently identify authors of such posts. In this paper, we explore how we can identify whether a given author's writing actually belongs to them by using various natural language and machine learning approaches.

Methods to determine authorship of a writing sample have been developed and studied a lot since $19^{th}$ Century. The most famous example was figuring out authors for federalist papers, a collection of politically-motivated articles. It has formed a branch in nature language processing called stylometry. There are several stylometric features we can consider in order to measure the similarities between author's writing and the target writing sample. These features can be primarily categorized as lexical, syntactic, semantic, and application specific. As technology advances and computing becomes faster and more accessible, various statistical and machine learning methods make it possible to use these features to distinguish between writing samples of various authors with relatively high confidence. In this paper, we study Reddit posts and are able to tell whether the post belongs to a given user or not. Reddit is a social news aggregation website where users can share news articles and make comments to contribute to discussion. Normally, a comment is short (no more than 100 words) and casual in nature. We are particularly interested in these comments to see which user account they belong to. We built various word-level and character-level models utilizing stylometric features and aggregated the results using an unweighted majority voting system to arrive at a final decision. With this method of ensemble learning, we are able to achieve fairly high system performance.

## 2 Motivation

We are in an information era now where social media is booming and a ton of data is generated daily. There are several applications of being able to identify the author of a short text, such as detecting hacked social media accounts. If we are able to verify that posts coming from a certain account do not seem to be written by the actual owner of the account, that could be a possible indication of a hacked account. Moreover, we could establish the credibility of a source. We figure out the credibility of a piece of content (post or a comment) written by a user on social media if we are able to detect whether the content posted is actually by the account owner or someone else. This is very useful as many notable politicians, such as presidents, are using social media to post politically motivated writings. In addition, we can prevent negative phenomenon like bullying in group texts, messages, or an aggregated piece of text if we are able to identify who the author might be from a potential pool of contributors.

## 2.1 Personal Motivation

One of our close friends had his Facebook account hacked multiple times and did not realize that he had been hacked until a few days had passed even though there were posts made from his account. When we noticed that the content posted from his account did not look like something he would write, we notified him immediately, and he was able to regain control of his account.

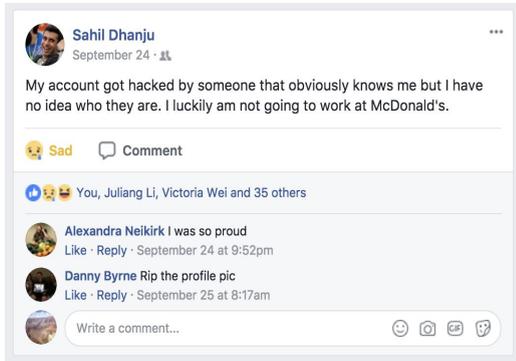

**Figure 1:** Our friend's post after getting hacked

This led to us exploring more about what we could do to aid this process of detecting authorship in social media posts and its various use cases.

## 3 Result

We collected over 1000 comments made by Redditor VRCkid and 1000 random comments made by top 20 Reddit users. We built five models for VRCkid and partitioned 70 percent of his posts as training data and cross-tested with 30 percent of VRCkid's remaining posts combined with 30% of other aggregated random Redditors' posts. We used 10-fold cross-validation and were able to achieve a precision, recall and F-measure of 0.82, 0.926 and 0.869 respectively.

## 4 Experiment

We began by collecting user comments from Reddit and applied a combination of five different models to conduct our experiment.

## 4.1 Data Collection

We used Python Reddit API Wrapper (PRAW) library to pull user data from Reddit and save it in JSON format so we can interpret and process it easily later. We considered bias in social media posts on Reddit and realized that most posts made by authors in subreddits were a collection of best practices, questions, and links, and these did not accurately describe the author's natural writing style. Hence, we decided to focus on the authors' comments instead.

Comments on Reddit seemed to more naturally reflect author's writing and provided a better sense of their writing style. We gathered 1000 random comments for the 20 top Redditor's in addition to all of VRCkid's comments, and we separated them by user names. Each file we stored represented a user and contained an array of strings.

## 4.2 Method 1: Word Frequency

A comment can be viewed as an array of word tokens. The tokens can be the actual word, a punctuation or a number. A natural way to detect if the person is the actual writer of the text is to check the vocabulary richness of this person's writing against the vocabulary richness of the text [1]. This approach makes sense since we do not expect the author of the text to suddenly use a lot of different words than what they would normally use in their other writings. Hence, we want to build a vector of word frequencies and pick the top results [2]. To get word tokens, we use NLTK library's casual tweet Tokenizer. It also works well in extracting the word tokens for our Reddit comments since the comments on Reddit are similar to tweets on Twitter in that they are both short, casual and may contain online slangs or grammatical errors.

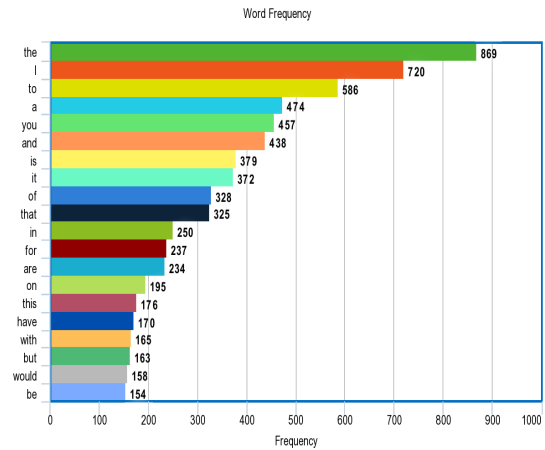

**Figure 2:** Top 20 words for VRCkid

The above graph shows the top twenty most frequently used words by Redditor VRCKid. Even though these words are fairly common and not particularly interesting, we did find some unique words that only VRCkid would use in the top 400 most frequent words. For example, "Google", "game" and "Dota" appear fairly often as this person is a gamer with interests in technology. With that information, we can build a bag of words representation of the text in order to carry out the machine learning algorithm. We decided to use a maximum entropy algorithm to train the model. We trained the model using 70% of VRCkid's comments and 70% of random Redditors' comments, making the total training set a matrix with a size equal to N * 1400, where N is the feature size and 1400 is the number of training samples. The feature size directly depends on the size of the bag of words. We tried many size of features in order to see which size would produce best results.

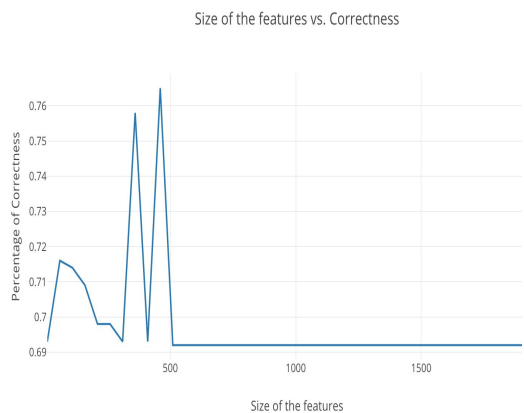

**Figure 3:** Graph comparing accuracy with feature size for bag of words

As can be seen in the above graph, the performance of the algorithm only minorly increases as the size of features increase. And once the size increases to 500, the correctness goes down and stays at 69 percent. The highest performance, which is 76.5 percent, occurs when feature size equals 480. It is interesting to see that the performance actually decreases with more features. It is also worthwhile to note that VRCKid has about 5065 unique words in our training sample. We also tried increasing the feature size and using a support vector machine (SVM) algorithm to train the model since maxent is known for overfitting and SVM can deal with higher dimensionality training data better. However, the accuracy we get with 5000 features is almost the same as 1000 features. The drawback of this approach is that it completely disregards the word order. For example, the phrase, "Wake up", is seen as "Wake" and "Up", which does not mean the same thing when it is separated. To solve this, we can use word N-gram to include the relative order of the words. However, since there is almost an unlimited combination of words, the resulting data is very sparse so it does not produce better results than a single word frequency.

### 4.3 Method 2: Character N-gram Frequency

Similar to the word frequency approach, this approach views the comments as a sequence of characters [3]. Viewing comments this way enables us to count character and punctuation frequencies. However, if we only count frequency of characters, the dataset would be very limited as there are only 26 letters in English. Therefore, we count the character n-grams in the text. This enables us to capture contextual information as well as lexical information. Another benefit of doing character-base n-gram is that the noise in the data does not affect the training sample as there are only finite set of n-gram candidates. Assuming the data is only in ASCII form, if we do unigrams, the maximum size is 128. If we use bigrams, the maximum size would be 16,384. The size does not grow much in comparison to the bigram for counting words. When considering unigrams, the most frequently occurring character is " ", the empty space character followed by "e". When considering bigrams, the most frequently occurring pair is "e" and " " followed by " " and "t". With the frequency counts, we are able to extract top elements and use a bag of words method to decide the features and how to construct them; this is very similar to what we did in the previous method. There are many ways to pick the number of grams and the size of the data. We tried a few combinations, trained them using logistic regression and were able to produce the following results as mentioned in the table.

| Features Size / Gram-size | 100 | 200 | 300 | 400 | 500 |
|---|---|---|---|---|---|
| 1 | 0.763 | 0.763 | 0.764 | 0.766 | 0.763 |
| 2 | 0.728 | 0.73 | 0.741 | 0.772 | 0.772 |
| 3 | 0.722 | 0.731 | 0.732 | 0.729 | 0.728 |
| 4 | 0.699 | 0.689 | 0.691 | 0.730 | 0.741 |
| 5 | 0.694 | 0.696 | 0.728 | 0.718 | 0.729 |
| 6 | 0.684 | 0.694 | 0.687 | 0.707 | 0.697 |
| 7 | 0.689 | 0.681 | 0.688 | 0.681 | 0.692 |
| 8 | 0.691 | 0.684 | 0.674 | 0.695 | 0.698 |
| 9 | 0.703 | 0.694 | 0.688 | 0.686 | 0.684 |
| 10 | 0.702 | 0.699 | 0.696 | 0.691 | 0.690 |

**Table 1:** Accuracy per feature and n-gram size for character n-grams

As shown in the table, unigram performs very well and is fairly stable, since the size of features don't impact the result as much. This can be explained by the fact that the maximum size of unigram is fairly small. On the other hand, as the gram size increases, the accuracy actually decreases. This can be explained by the fact that data would be much sparser with more grams. However, bigram produces the best result and it looks like there is an upward trend. Therefore, we decided to focus on bigrams and did some additional experiments on bigram.

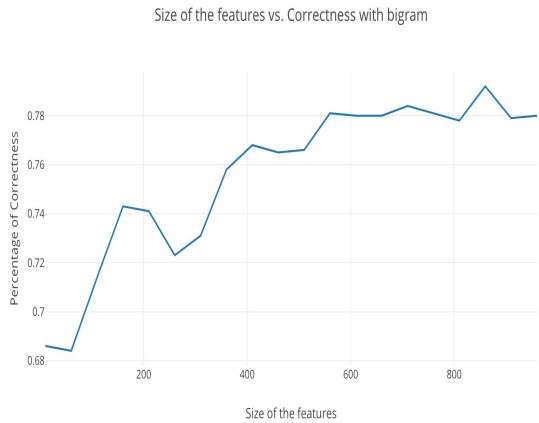

**Figure 4:** Graph comparing accuracy with feature size for bigrams

As can be seen from the above graph, the performance of the algorithm generally increases as the size of features increases. The correctness goes up rapidly from when the feature size is 10 to when the feature size increases to 400. However, it seems to have a ceiling, and the highest performance of 79 percent occurs when the feature size equals 860. Overall, this approach produces very good results.

### 4.4 Method 3: Parts of Speech

Another approach to analyze the given comment is to look at the syntactic information. The author of the text may subconsciously write sentences using similar syntactic structure [2]. We can use this information to generate a fingerprint of the author. One important aspect of syntactic information is parts of speech. There are eight major parts of speech: Noun, Pronoun, Verb, Adverb, Adjective, Conjunction, Preposition and Interjection. With those parts of speech, we can consider n-grams to count the combinations occurring in the comment, similar to what we did with method 2. Also, similar to method 2, there are various ways to pick the gram size and feature size for training. Therefore, we tried a few different combinations, trained them using logistic regression and were able to produce the following results as mentioned in the table.

| Gram size | Top 1 frequency | Top 2 frequency | Top 3 frequency |
|---|---|---|---|
| 1 | NN | IN | DT |
| 2 | DT, NN | IN, DT | PRP, VBP |
| 3 | IN, DT, NN | DT, JJ, NN | DT, NN, IN |
| 4 | NN, IN, DT, NN | IN, DT, NN | IN, DT, JJ, NN |

**Table 2:** Most frequent parts of speech per n-gram size

As can be seen from the table, with unigram, the author VRCKid uses common nouns the most. With bigram, he uses determiners followed by common pronouns often. With trigram, he uses phrases such as "for good games" most frequently. With this information, we decided to try out a few combinations of gram size and the feature size to see how it performs.

| Gram size / Feature size | 1 | 2 | 3 | 4 | 5 |
|---|---|---|---|---|---|
| 30 | 0.729 | 0.681 | 0.689 | 0.691 | 0.689 |
| 80 | 0.732 | 0.698 | 0.675 | 0.691 | 0.686 |
| 130 | 0.714 | 0.676 | 0.678 | 0.694 | 0.682 |
| 180 | 0.725 | 0.694 | 0.679 | 0.675 | 0.685 |
| 230 | 0.725 | 0.722 | 0.694 | 0.679 | 0.680 |
| 280 | 0.725 | 0.719 | 0.692 | 0.685 | 0.674 |
| 330 | 0.738 | 0.704 | 0.698 | 0.676 | 0.685 |
| 380 | 0.725 | 0.716 | 0.696 | 0.690 | 0.684 |
| 430 | 0.724 | 0.720 | 0.701 | 0.678 | 0.680 |
| 480 | 0.727 | 0.716 | 0.699 | 0.690 | 0.680 |
| 530 | 0.719 | 0.718 | 0.685 | 0.679 | 0.682 |
| 580 | 0.720 | 0.717 | 0.700 | 0.688 | 0.682 |

**Table 3:** Accuracy per feature and n-gram size for parts of speech

Interestingly, unigrams produce the best performance, and as the size of the features grows, accuracy grows too, but there is a ceiling. We think that the reason multi-grams cannot compete with unigrams is because of data sparsity, similar to what method 2 suffers from. Overall, we are satisfied with this performance.

### 4.5 Method 4: Lexical K-Means Cluster

Unlike the previous three methods, this approach focuses on unsupervised learning where we analyze writing patterns and draw inferences without providing labeled data. We do cluster analysis using the very popular k-means clustering algorithm. In this case, the k value is two and we partition our data into two distinct clusters based on distance to the centroid of a cluster. This two cluster approach acts as a classifier since it groups the given data into texts authored by the given author and texts authored by everyone else based on our feature set. Our goal while constructing the feature set and picking features was to focus on features that captured distinctive aspects of an author's writing style and maintained consistency, even when the author wrote on different subjects on various forms of social media. This led us to explore lexical, structural, syntactic and punctuation features. We considered the average number of sentences per word, sentence length variation and lexical vocabulary diversity. Sentence variation is calculated by measuring standard deviation of the words per sentence. Sentence variation provides us with a measure of how much and how frequently the author varies their sentence length. Lexical vocabulary diversity represents the richness of the author's vocabulary and is based on how frequently the author repeats words in sentences as compared to using new words. For punctuation, we calculated the average number of most commonly used punctuation marks, like colon, semicolon and commas. These lexical and punctuation features capture distinctive aspects of the author's writing style while being topic agnostic. Below are some statistics we garnered from VRCkids's comments. The data represented in the graph below has been averaged for all sentences in comments.

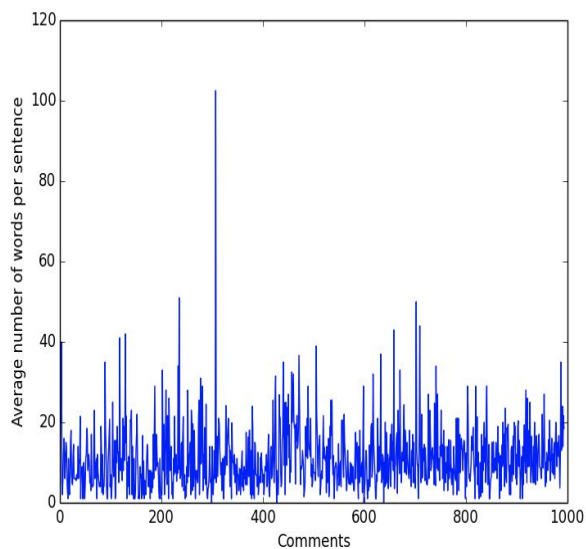

**Figure 5:** Graph of average number of words per sentence averaged across a comment for all comments

As we notice from the graph above, the average number of words per sentence for a comment by VRCkid is concentrated in the range from four to twenty words per sentence, with very few exceptions. This is a potentially important and discriminatory feature to consider since any comment with more than 25 words per sentence has a very low probability of being authored by VRCkid. We further analyzed other discriminatory lexical and punctuation features for VRCkid, and our results are presented in the tables below.

| Average number of words per sentence | 10.98 |
|---|---|
| Average sentence length variation | 2.55 |
| Average vocabulary diversity | 0.90 |

**Table 4:** Lexical diversity of comments

| Average semicolon per sentence | 0 |
|---|---|
| Average colon per sentence | 0.058 |
| Average commas per sentence | 0.19 |

**Table 5:** Punctuation diversity of comments

As we notice, VRCkid never used semicolons in his comments, and any comment with a semicolon has a very negligible probability of being authored by VRCkid. Overall, this model was able to achieve an accuracy of 69 percent individually using the feature set we specified above.

## 4.6 Method 5: Short Text Verification classifier

So far, some of the approaches we have considered rely on stylometric techniques that analyze linguistic styles and writing characteristics. Although stylometric techniques work well for author detection for large documents, they may perform poorly for short pieces of text because of their lack of structure and short length. To deal with this problem, we looked at a new supervised learning algorithm with an n-gram analysis approach that checks the identity of an author for a short text [4]. Stylistics features can be broadly classified as lexical, structural, syntactic and content specific. Within lexical features, n-gram features are effective and noise tolerant. While most previous approaches consist of looking at n-gram frequency, this approach looks at the presence or absence of n-grams and the relationship they have with the training dataset. This reduces the n-gram features to one, therefore leading to a reduction in complexity and the time consumed by the classifier for processing data. This model generates separate profiles for each user in the collection, and this is broken into a training and verification step. The training phase entails building the user profile, while the verification phase checks and verifies the user profile built through the training phase. The training phase is further divided into two parts - computing a user profile and computing a threshold. The first part of the training phase computes the user profile by extracting n-grams from the provided sample document while the second part computes a user threshold that we use later during the verification phase.

We begin by dividing the training data about a user into two subsets. We calculate the number of n-grams in the first subset, while we divide the second subset further into multiple blocks of characters of equal size. Next, we explore the percentage of unique n-grams shared by character blocks of other users with the set of unique n-grams occurring in the first split of our user's training set ($P_U^B$). The block is considered to be our user's work if and only if this calculated percentage of unique n-grams shared by the block with n-grams in the training set is greater than the sum of a specific computed user threshold and a predefined constant. Below is the algorithm used to calculate threshold for a given user.

```
/* U a user for whom the threshold is
   being calculated              */
/* I_1,...,I_m: a set of other users
   (I_k ≠ U)                     */
/* ε_U: threshold computed for user U
                                 */
Input: Training data for U, I_1, ..., I_m
Output: ε_U
1  begin
2    up ← false;
3    down ← false;
4    δ ← 1;
5    ε_U ← μ_U − (σ_U/2);
6    γ ← 0;
7    while δ > 0.0001 do
        /* Calculating FAR and FRR for
           user U                */
8       FRR_U, FAR_U =
        calculate(U, I_1, ..., I_m, ε_U, γ);
        /* Minimizing the difference
           between FAR and FRR   */
9       if (FRR_U − FAR_U) > 0 then
10          down ← true;
11          ε_U ← ε_U − δ;
12      end
13      if (FAR_U − FRR_U) > 0 then
14          up ← true;
15          ε_U ← ε_U + δ;
16      end
17      if (up & down) then
18          up ← false;
19          down ← false;
20          δ ← δ/10;
21      end
22    end
23    return ε_U;
24 end
```
**Algorithm 1:** Threshold calculation for a given user.

Reference: Brocardo et. al. [4]

We also consider FRR and FAR while iteratively computing a threshold for a user.

FRR - False Rejection Rate is the probability that the model would not recognize the correct author of the document.

FAR - False Acceptance Rate is the probability that the model will wrongly recognize someone as the author of the text.

The threshold is calculated through a supervised learning technique as depicted above, and is incrementally varied by minimizing the difference between FAR and FRR values, with a goal of making them equal. Once we have a threshold value, FAR and FRR are calculated as follows.

FRR calculation

```
Loop number_of_blocks times {
    if P_U^B < threshold + constant
        then false_rejection += 1
}
```
$false\_rejection\_rate = \frac{false\_rejection}{number\_of\_blocks}$

FAR calculation

```
Loop number_of_other_users times
{
    Loop number_of_blocks times
{
  if P_U^B ≥ threshold+constant
      then false_acceptance += 1
    }
}
```

$N^B$ = number_of_blocks
$N^{OU}$ = number_of_other_users
$FAR = \frac{false\_acceptance}{N^B * N^{OU}}$

We evaluated this approach on VRCkid's data and varied our parameters to compare results. We began by varying the predefined constant gamma ($\gamma$) that is used along with the threshold to verify the author of text.

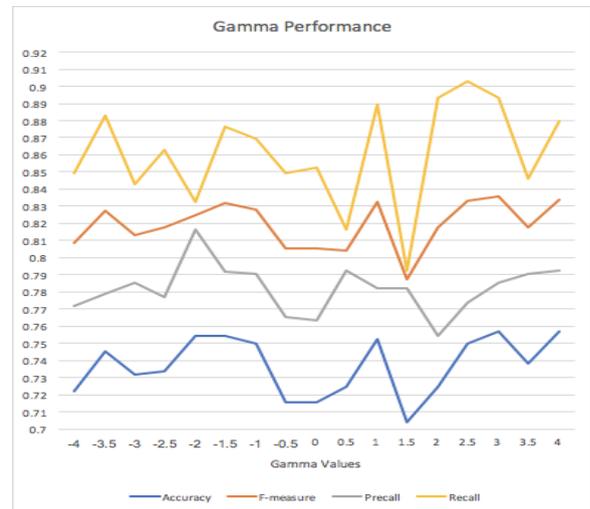

**Figure 6:** Graph comparing performance with Gamma

The performance (F-measure and accuracy) peaks when $\gamma = 3$ for our given training and test set for

VRCkid. In addition, we also looked at variation from different values for n-grams.

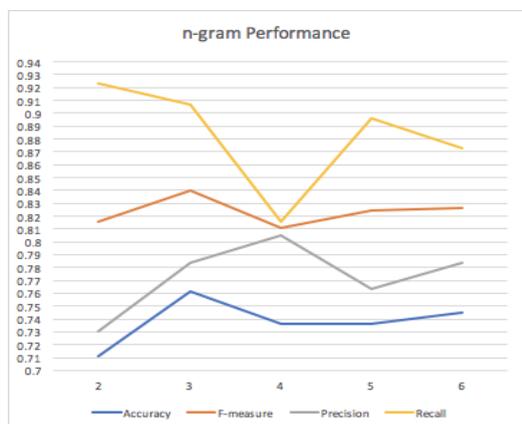

**Figure 7:** Graph comparing performance with n-gram size

Similar to $\gamma$, the n-gram performance also peaks when $n = 3$. Overall, we were able to get a 77 percent accuracy with this model.

### 4.7 The Voting System

We have discussed five different classifiers to train and detect whether the text belongs to the author or not. Every classifier has its strengths and weaknesses. For example, word frequency classifier cannot capture the natural ordering of the words, and the n-gram approach suffers from data sparsity. Therefore, we want to combine their strengths to overcome their weaknesses. The structure of the final classifier is very similar to that of a neural network; there is only one layer with five nodes. The final layer uses an unweighted majority-vote strategy to determine the output based on the output from the middle layer. For example, if the output vector for the middle layer is $[1, 1, 1, 0, 0]$, we would output true, as there are more ones that outweigh the zeros. We output false if it is the other way round. With this strategy, we are able to get even better results than what the classifiers could do individually. The following are accuracy results from a 10-fold cross-validation test.

| Fold 1 | 0.819 |
|---|---|
| Fold 2 | 0.810 |
| Fold 3 | 0.824 |
| Fold 4 | 0.819 |
| Fold 5 | 0.794 |
| Fold 6 | 0.813 |
| Fold 7 | 0.794 |
| Fold 8 | 0.789 |
| Fold 9 | 0.826 |
| Fold 10 | 0.778 |
| Average | 0.807 |

**Table 6:** Accuracy attained by an unweighted majority vote

The performance is slightly better than the best performance made by character base n-gram. This way of classification is called ensemble learning, a machine learning paradigm where multiple models are trained to tackle the same problem. As we can see, ensemble learning yielded better results for classifying short social media text messages.

## 5 Conclusion

We discuss five approaches to tackle the problem of figuring the authorship of a short text, namely Reddit comments in our case. The five approaches explore the relationship and limitations in lexical feature, character based features and syntactic features of the data. In the end, we are able to combine the classifiers to create a voting system and attain a precision, recall and F-measure of 0.82, 0.926 and 0.869 respectively.

## 6 Limitations

Reddit is one source we looked at. Although, we could apply our approach to other forms of social media, it would be harder to do so for social media websites like Facebook, where users compose even shorter text. Moreover, we did not take into consideration other factors that could lead to identifying authors, like the location the post was made at. Lastly, our targeted Redditor, VRCkid, only has less than one thousand comments. The data may not be enough to generate a very accurate fingerprint for authorship detection.

## 7 Future Work

There are many other stylometric features that can be explored further. We have only touched the surface of lexical analysis via word frequency, character analysis via character n-grams and syntactic structure via part of speech parsing. In addition to these features, we can also try out concepts like semantic measurement. For example, does the author always use words that are synonyms to what they normally use? Also, sentence and phrase structure can be explored by using a partial or full parser to break down the text. In addition, our voting system is unweighted. We could potentially explore adding weights to our final majority vote vector to improve performance of our system.